\documentclass[11pt]{article}

\usepackage[final]{acl}

\usepackage{times}
\usepackage{latexsym}

\usepackage[T1]{fontenc}

\usepackage[utf8]{inputenc}

\usepackage{microtype}

\usepackage{inconsolata}

\usepackage{graphicx}

\usepackage{multirow}
\usepackage{booktabs}
\usepackage{graphicx}
\usepackage{subcaption}
\usepackage{fvextra}

\title{A Multistage Extraction Pipeline for Long Scanned Financial Documents: An Empirical Study in Industrial KYC Workflows}

\author{
 \textbf{Yuxuan Han},
 \textbf{Yuanxing Zhang},
 \textbf{Yushuo Wang}, 
 \textbf{Yichao Jin},
 \textbf{Kenneth Zhu Ke},
 \textbf{Jingyuan Zhao}
\\
\texttt{\{yuxuanhan, yuanxingzhang, yushuowang, jinyichao, kennethzhu, jingyuanzhao\}@ocbc.com}
\\
 \texttt{OCBC, Singapore}
}

\begin{document}
\maketitle
\begin{abstract}
Structured information extraction from long, multilingual scanned financial documents is a core requirement in industrial KYC and compliance workflows. These documents are typically non-machine-readable, noisy, and visually heterogeneous. They usually span dozens of pages while containing only sparse task-relevant information. Although recent vision–language models (VLMs) achieve strong benchmark performance, directly applying them end-to-end to full financial reports often leads to unreliable extraction under real-world conditions.

We present a multistage extraction framework that integrates image preprocessing, multilingual OCR, hybrid page-level retrieval, and compact VLM-based structured extraction. The design separates page localization from multimodal reasoning, enabling more accurate extraction from complex multi-page documents.

We evaluated the framework on 120 production KYC documents comprising about 3000 multilingual scanned pages. Across multiple OCR–VLM combinations, the proposed pipeline consistently outperforms direct PDF-to-VLM baselines, improving field-level accuracy by up to \textbf{31.9 percentage points}. The best configuration, PaddleOCR with MiniCPM-o-2.6, achieves \textbf{87.27\%} accuracy. Ablation studies show that page-level retrieval is the dominant factor in performance improvements, particularly for complex financial statements and non-English documents.
\end{abstract}

\section{Introduction}

Information extraction from unstructured documents is a critical step in many operational processes at international financial institutions, including Know-Your-Customer (KYC) onboarding, anti-money-laundering (AML), and regulatory compliance workflows. These processes rely on large volumes of customer-submitted materials such as audit reports, financial statements, payslips, bank statements, and identity documents. To improve efficiency and reduce manual effort, institutions increasingly deploy automated document understanding systems to extract structured information from such submissions.

Despite its importance, accurate information extraction in this setting remains challenging. Most documents are non-machine-readable scans of varying quality, often affected by low resolution, skew, compression artifacts, and background noise. Layouts are heterogeneous, interleaving narrative text, tables, charts, stamps, and handwritten annotations. Financial documents also contain domain-specific terminology and multilingual content, complicating normalization and extraction. 

Recent advances in large Vision–Language Models (VLMs) enable joint reasoning over visual and textual inputs and show strong performance on document benchmarks. Yet directly applying such models to real-world financial documents is often impractical. The length and complexity of scanned files substantially increase computational cost and can degrade extraction reliability under end-to-end processing. While visual retrieval-augmented approaches such as ColPali \citep{faysse2025colpali} and VisRAG \citep{yu2025visrag} improve document-level retrieval, they remain significantly more expensive than text-based inference \citep{rajendran-etal-2025-ecodoc}, limiting scalability in high-volume workflows.

In this paper, we propose a multistage pipeline that integrates image preprocessing, multilingual OCR, page-level retrieval, finance-specific prompt adaptation, and compact VLM-based extraction. This modular design reduces computational cost while improving extraction accuracy, enabling efficient processing of long, heterogeneous document collections. Using an internal corpus of real-world KYC documents, we conduct a comprehensive empirical study evaluating accuracy, efficiency, and robustness. The proposed framework achieves up to 31.9 percentage points higher accuracy than directly applying VLMs to entire documents, while maintaining comparable service latency. These results provide practical guidance for building reliable extraction systems for long scanned documents in industry settings.
\section{Related Works}
Structured extraction from financial reports builds upon OCR, document layout understanding, vision–language modeling, and financial NLP. Progress across these areas has improved the integration of textual, visual, and structural cues, achieving strong results on benchmark datasets.

\textbf{OCR and Image Preprocessing} underpin scanned document understanding by converting images into machine-readable text and mitigating noise. Classical engines such as Tesseract \citep{smith2007overview} remain widely used, while deep learning–based OCR systems improve robustness on multilingual and low-quality scans \citep{cui2025paddleocr3, easyocr2020}. Complementary preprocessing techniques—including page segmentation \cite{chen2017convolutional}, skew correction \cite{akhter2020improving}, and image enhancement \cite{anvari2021survey}—further enhance OCR reliability under real-world conditions.

\textbf{Document Layout Understanding and Vision–Language Models (VLMs)} Layout-aware models such as the LayoutLM family \citep{xu2020layoutlm, xu2021layoutlmv2, huang2022layoutlmv3} incorporate spatial features to improve extraction from visually rich documents. Subsequent work decouples text and layout for language-independent understanding \citep{wang2022lilt}, or integrates layout signals into LLMs without heavy image encoders \citep{li2023docllm}. OCR-free approaches, including Donut \cite{kim2022donut} and UReader \cite{ye2023ureader}, explore end-to-end modeling from document images to structured outputs. More recent VLMs such as mPLUG-DocOwl and DocOwl2 \cite{hu2024mplug, hu2025mplug} extend these directions with unified structure learning and high-resolution representations for complex, multi-page documents.

\textbf{Evolution of Financial Information Extraction} Early financial extraction methods relied on rule-based systems \citep{sheikh2012rule, im2015rule}, encoding domain knowledge but lacking flexibility. Later work adopted multimodal approaches combining text, tables, and figures \citep{chen2021finqa, singh2024finqapt}. Recent studies leverage VLMs to process complex financial layouts, including fine-tuning models to convert tables into structured text \citep{tan2025finetuning, poznanski2025olmocr}, generating intermediate structured representations for improved numerical reasoning \citep{srivastava2025enhancing}, and incorporating explicit layout modalities to enhance extraction accuracy \citep{aida2025enhancing}.

Despite strong benchmark performance, many approaches are evaluated on simplified or synthetic datasets \citep{bradley2026synfintabs} that do not reflect the complexity of real-world financial documents. In practice, reports are often scanned, multi-page, and visually heterogeneous, posing challenges for end-to-end modeling \citep{tan2025finetuning}. Moreover, the high computational cost of large VLMs limits scalability in high-volume workflows. Although compact models demonstrate promising efficiency–accuracy tradeoffs, systematic evaluation of cost-aware extraction pipelines on long, noisy financial documents remains limited. Our work addresses this gap through a multistage framework for scalable KYC extraction, accompanied by a comprehensive empirical study to quantify the impact of each pipeline component on performance and efficiency.
\section{Problem Statement and Methodology}


\subsection{Problem Statement}
We consider the task of structured information extraction from real-world financial documents in Know-Your-Customer (KYC) workflows. Given a financial document $D = \{p_1, p_2, \ldots, p_n\}$ consisting of n scanned pages, the goal is to extract a predefined set of target fields $F = \{f_1, f_2, \ldots, f_m\}$, where each field corresponds to a financial attribute required for customer due diligence, such as revenue, net profit, or dividends. 

This task is challenging for three main reasons. First, scanned documents are affected by noise such as low resolution, skew, and compression artifacts. Second, financial reports frequently contain complex layouts that combine narrative text, tables, and mixed text–table regions. Third, although documents are long, only a small subset of pages is relevant to any given target field, making end-to-end processing inefficient.

Accordingly, we evaluate extraction systems along two key dimensions. \textbf{Accuracy} measures robustness to OCR noise, layout variability, and multilingual content. \textbf{Efficiency} is assessed in terms of per document service latency. Our objective is preserve end-to-end latency comparable to direct PDF-to-VLM baselines while achieving substantially stronger extraction performance on real-world KYC financial documents.

\subsection{Proposed Multistage Pipeline Overview}

\begin{figure*}[t]
\centering
\includegraphics[width=\linewidth]{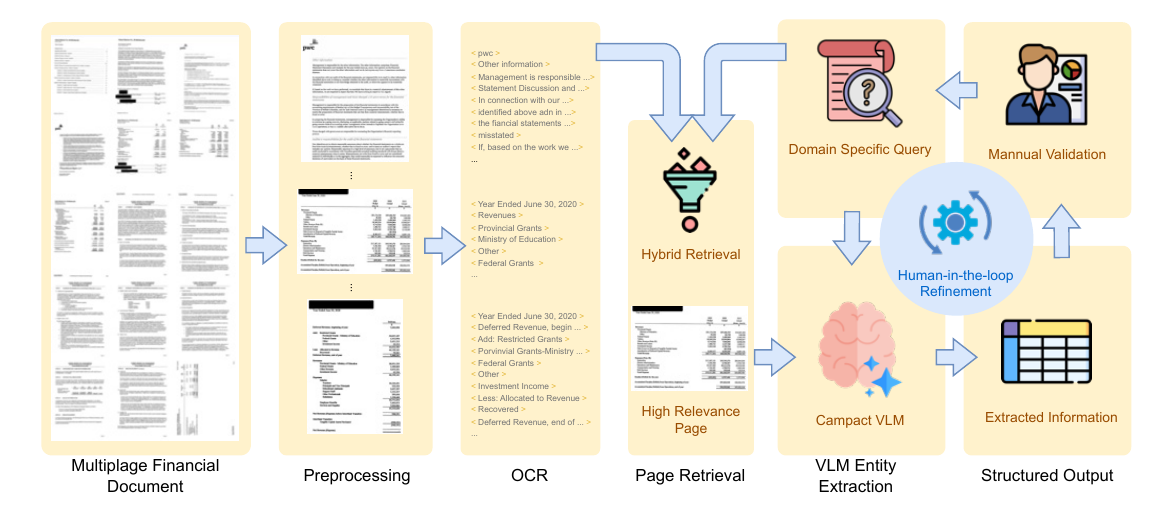}
\caption{Overview of the proposed multi-stage financial document extraction pipeline.}
\label{fig:pipeline}
\end{figure*}

To address the accuracy and efficiency challenges of KYC document processing, we adopt a multistage extraction pipeline that reserves expensive vision–language inference for only those pages likely to contain relevant information, rather than processing entire documents end-to-end.

As illustrated in Figure \ref{fig:pipeline}, each document is first split into individual pages and processed by image preprocessing and multilingual OCR to recover textual content. Based on the OCR outputs, a page-level retrieval component ranks pages by their relevance to target financial fields and filters out most irrelevant content. The remaining pages are then processed by compact vision–language models for fine-grained extraction through joint text–layout reasoning. In high-stakes KYC settings, extracted fields can optionally be reviewed through a human-in-the-loop process for efficient verification and prompt refinement.

By decoupling preprocessing, retrieval, and extraction, the proposed pipeline enables scalable processing of long, heterogeneous financial documents under strict computational constraints.

\subsection{Image Preprocessing}
In KYC workflows, financial documents often suffer from low resolution, skew, background noise, and highly variable layouts. These make robust image pre-processing essential for generating accurate OCR results. This stage converts noisy scanned pages into clean and reliable inputs sequence of preprocessing operations. 

\textbf{Segmentation} We use OpenCV’s edge detection and contour-finding algorithms \citep{xie2013image} to identify content-bearing regions and remove large blank areas and extraneous borders. This step improves the visibility of small text and tables, allowing subsequent processing to focus on relevant regions of the page.

\textbf{Skew and rotation correction} We use two-stage correction to address common scanning misalignments that can significantly affect OCR accuracy. We first apply PaddleOCR’s document orientation classifier \citep{cui2021pp} to correct coarse rotations, followed by fine-grained skew correction using the Hough transform \citep{ahmad2021efficient} to align text lines horizontally. This two-stage approach ensures robustness against both gross rotation errors and subtle scanning distortions.

\textbf{Re-normalization} Cropped pages are rescaled using Bicubic interpolation \citep{han2013comparison} to preserve the sharpness of small characters and table borders. We further apply Contrast Limited Adaptive Histogram Equalization (CLAHE) \cite{reza2004realization} for local contrast normalization, together with light Gaussian denoising to suppress background artifacts such as stains or shadows.

\subsection{OCR and Page Retrieval}
After pre-processing, a multilingual OCR engine is applied to transcribe textual content from each page. We explore PaddleOCRv3 \citep{cui2025paddleocr3} and EasyOCR \citep{easyocr2020} for their efficiency and broad language coverage, as well as the proper support for both printed and handwritten text commonly encountered in international KYC documents. The transcribed texts serve as the basis for subsequent page-level retrieval and compact VLM-based extraction.

Using the OCR output, we perform page-level retrieval to identify pages relevant to specific KYC fields. For each target field, we construct a predefined query combining three components: (i) domain-specific financial terms (e.g., revenue, net profit, dividends), (ii) document-type cues indicating typical locations of such information (e.g., financial statements, audit reports, payslips), and (iii) language-specific keywords to support multilingual submissions. This structured query design enables adaptation across extraction tasks and document types without retraining.

To enhance robustness, we adopt a hybrid retrieval strategy integrating lexical and semantic matching. BM25 \citep{robertson2009probabilistic} captures exact keyword matches and term-frequency signals effective in noisy OCR text, while a sentence embedding model computes dense representations for semantic similarity \citep{reimers2019sentence}, handling paraphrases and non-standard terminology. The final relevance score for each page is obtained by combining the lexical and semantic scores, balancing the precision of keyword-based retrieval with the recall of semantic similarity.

This hybrid OCR-based retrieval stage significantly reduces the number of pages forwarded to the VLM while maintaining robustness to OCR errors, multilingual variation, and heterogeneous layouts. As a result, computationally intensive extraction is applied only to a small subset of high-relevance pages, enabling scalable processing of long financial documents in KYC workflows.

\subsection{Extraction Using Compact VLMs and Human-in-the-Loop Improvement}
After page-level retrieval, only a subset of high-relevance pages is forwarded to the extraction stage. Instead of applying large VLMs to entire documents, we use compact VLMs to perform structured extraction on the filtered pages, reducing computational cost while retaining multimodal reasoning over text and layout.

For each target financial field, extraction is guided by a structured prompt that extends beyond retrieval queries. In addition to domain-specific keywords and document-type cues, the prompt includes output format instructions to standardize downstream processing. 

To support operational validation, the model output includes an additional remarked field. Alongside each extracted value, the VLM may provide brief comments when potential ambiguity is detected. These remarks provide contextual information to assist reviewers during verification.

In KYC workflows, all extracted fields are subject to manual review prior to downstream decision-making. During validation, extracted values need to be confirmed and/or corrected by analysts. This manual review is a regulatory requirement in production KYC workflows, rather than an additional cost introduced by our pipeline, and should be viewed as an existing operational overhead. Observed correction patterns are subsequently used to refine field-specific prompts and retrieval queries, enabling iterative improvement. In this sense, the human-in-the-loop component repurposes this mandatory review step for prompt refinement, adding no meaningful extra burden. For fairness in evaluation, analyst corrections are not incorporated into the reported accuracy metrics, which reflect the raw system outputs prior to manual intervention.

By combining compact VLM-based extraction with structured output constraints and systematic validation, the framework supports efficient and reliable structured information extraction in industrial KYC settings for financial institutions.
\section{Experiments}

This section presents a comprehensive evaluation of the proposed multistage extraction framework. We evaluate the system on a real-world, multilingual corpus of long, scanned financial documents and analyze extraction accuracy over various experiment setups. Beyond comparing different model backbones, we conduct systematic ablation studies to quantify the contribution of each component.

\subsection{Dataset}

The evaluation dataset consists of 120 real-world financial documents collected from production KYC workflows, including financial audit reports and employee payslips used for customer due diligence and income verification. In total, the dataset contains about 3000 scanned pages. Document lengths vary substantially, ranging from 1–3 page payslips to audit reports exceeding 80 pages.

The documents are in multiple languages, containing English, Indonesia Bahasa, Simplified and Traditional Chinese. All the files are non-machine-readable scanned documents and exhibit real-world artifacts such as skew, low resolution, compression noise, stamps, and heterogeneous layouts. Many pages combine narrative text, dense financial tables, and semi-structured disclosures within the same document, making structured extraction particularly challenging.

Each document type is associated with a predefined set of target financial fields aligned with its semantic structure. All fields are manually annotated and verified by domain analysts to ensure consistency and correctness. The dataset contains sensitive production KYC documents, and regulatory and privacy obligations prevent public release. Detailed dataset statistics and field definitions are provided in Appendix~\ref{sec:dataset}.

\subsection{Experimental Setup}

We evaluate the proposed framework across multiple OCR–VLM combinations and controlled pipeline variants to assess the effectiveness and robustness under real-world document conditions. Two multilingual OCR backbones, PaddleOCRv3 \citep{cui2025paddleocr3} and EasyOCR \citep{easyocr2020}, are used to examine the sensitivity to transcription quality. For visual–language reasoning, we compare MiniCPM-o-2.6 \cite{yao2024minicpm}, Gemma-3-27B-IT \citep{gemma_2025}, and Qwen3-VL-8B-Instruct \cite{bai2025qwen3}. All three models receive identical prompts and document images, with fixed limits on the number of input files and response tokens. Experiments are conducted on a single NVIDIA A100 80GB GPU with identical CPU and RAM allocation, while all other settings follow their default configurations.

For each OCR–VLM pair, we evaluate five variants, including, 1) the full pipeline; 2) removal of image preprocessing; 3) removal of page-level retrieval; 4) removal of language-adapted and finance-specific structured prompting; and 5) a direct PDF-to-VLM baseline without OCR or retrieval. Extraction performance is measured using field-level accuracy, where a prediction is correct only if it matches the normalized ground truth. The reported results are averaged across all fields and documents under a unified evaluation protocol.

\subsection{Overall Results}

  \begin{table*}[t]                                                                                            
  \centering
  \small                                                                                                       
  \setlength{\tabcolsep}{4pt}
  \begin{tabular}{llccccc}
  \toprule
  \textbf{OCR} & \textbf{VLM}
  & \textbf{Full}
  & \textbf{-ImgPrep}
  & \textbf{-Retrieval}
  & \textbf{-Prompt}
  & \textbf{Direct VLM} \\
  \midrule

  \multirow{3}{*}{PaddleOCR}
  & MiniCPM-o-2.6        & \underline{\textbf{87.27}} & 71.00 & 63.25 & 84.38 & 55.38 \\
  & Gemma-3-27b-it       & \textbf{72.97} & 60.89 & 52.10 & 72.57 & 47.64 \\
  & Qwen3-VL-8B-Instruct & \textbf{85.30} & 77.17 & 63.78 & 80.97 & 55.91 \\
  \midrule
  \multirow{2}{*}{EasyOCR}
  & MiniCPM-o-2.6        & \textbf{75.20} & 67.59 & 59.45 & 74.28 & 54.20 \\
  & Gemma-3-27b-it       & \textbf{65.09} & 58.92 & 48.29 & 64.30 & 47.51 \\
  \bottomrule

\end{tabular}
\caption{Field-level accuracy (\%) across OCR–VLM combinations and pipeline variants. The five configurations include the full multistage pipeline, removal of image preprocessing (-ImgPrep), removal of page-level retrieval (-Retrieval), removal of structured language-adapted prompting (-Prompt), and a direct PDF-to-VLM baseline without OCR or retrieval.}
\label{tab:main_results}
\end{table*}

Table \ref{tab:main_results} presents the overall extraction accuracy across all OCR–VLM combinations and pipeline variants. Across all tested pairs, the proposed multistage pipeline consistently achieves the highest accuracy. Conversely, directly feeding full PDFs into the VLM yields the lowest performance in every configuration, underscoring that \textbf{structured processing is essential for reliable data extraction} from long, scanned financial documents. The detailed analysis of pipeline effectiveness and module contributions follows in the subsequent sections.

The strongest overall result is achieved using PaddleOCR paired with MiniCPM-o-2.6, reaching \textbf{87.27\%} accuracy. In particular, MiniCPM-o-2.6 remains competitive across both OCR backbones and consistently outperforms the significantly larger Gemma-3-27b-it model under identical settings. This performance gap likely stems from the specialized architecture of MiniCPM-o-2.6, which uses adaptive high-resolution visual encoding and high-density OCR tokenization specifically optimized for document-centric tasks \cite{yao2024minicpm}, while Gemma-3-27b-it functions as a more general-purpose multimodal model.

To further validate the efficacy of our framework, we extended our evaluation to include Qwen3-VL-8B-Instruct \cite{bai2025qwen3}, the current state-of-the-art (SOTA) among open-weight vision-language models. Our results indicate that the proposed multistage pipeline consistently outperforms the direct PDF-to-VLM baseline by a substantial margin, confirming that our efficiency and accuracy gains persist even when integrated with leading SOTA architectures. While QWen3-VL-8B-Instruct exhibits strong performance, its absolute accuracy in this specific setting remains slightly below that of MiniCPM-o-2.6. We attribute this discrepancy to the fact that our prompt engineering was primarily calibrated for the MiniCPM deployment environment, rather than a lack of inherent model capability.

\subsection{Analysis}

\subsubsection{Effectiveness of the Multistage Pipeline}

\begin{figure}[t]
\centering
\includegraphics[width=\linewidth]{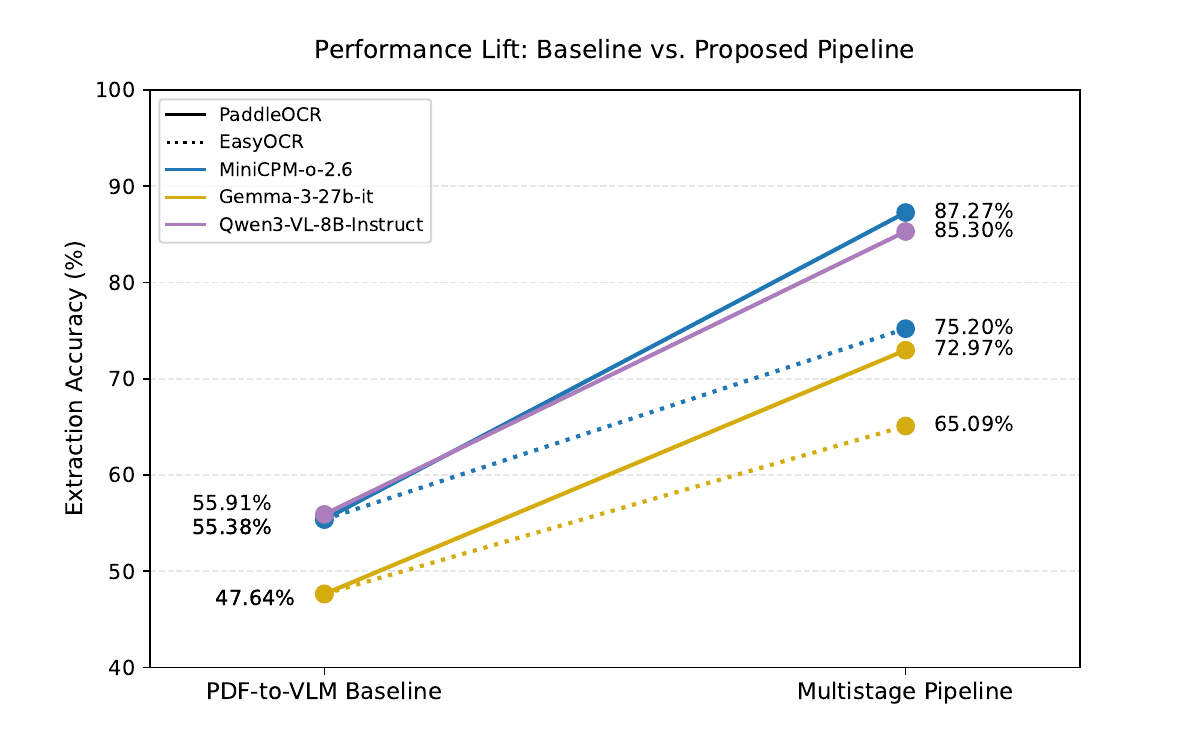}
\caption{Performance lift from PDF-to-VLM Baseline to Multistage Pipeline. The plot illustrates a universal accuracy improvement across all test settings.}
\label{fig:performance_lift}
\end{figure}

Figure \ref{fig:performance_lift} visualizes the performance trajectory of four distinct OCR–VLM configurations when transitioning from a direct PDF baseline to our proposed multistage pipeline. The steep positive slopes observed across all five lines indicate a universal performance increase, independent of the underlying model architectures.

Quantitatively, the pipeline delivers a substantial accuracy improvement ranging from \textbf{20.7\%} to \textbf{31.9\%} across all settings. For the specialized MiniCPM-o-2.6 model paired with PaddleOCR, the pipeline maximizes potential, increasing accuracy to a peak of \textbf{87.27\%}. Crucially, the system demonstrates the same efficacy for general-purpose Gemma-3-27b-it and QWen3-VL-8B-Instruct models, with performance improving by \textbf{25.33-29.39 percentage points}, corresponding to \textbf{over 50\%} relative gain over direct baseline.

Notably, this substantial accuracy improvement does not come at the cost of increased latency. The per-page inference time remains comparable to the direct baseline (Appendix \ref{subsec:latency}). Moreover, the retrieval stage reduces the average number of pages sent to the VLM by approximately 70\%, leading to lower token usage in practice. 

Overall, the proposed pipeline serves as a robust architectural layer that consistently unlocks the extraction capabilities of diverse multimodal systems in complex financial contexts while remaining computationally efficient and practically cost-effective.

\subsubsection{Module-wise Ablation Study}

We perform a controlled ablation by removing one module at a time and analyse the absolute accuracy drop in Table~\ref{tab:main_results}. This pattern is consistent in all OCR–VLM configurations, revealing a clear module importance hierarchy. Visualization of the trend is provided in Appendix \ref{sec:result_module}.

Page-level retrieval is the most critical component. Its removal causes a substantial accuracy decrease of 16.8–24.0 percentage, indicating that accurate localization of relevant pages is essential for effective reasoning in long financial documents.

Image preprocessing is the second most influential factor, with a performance drop of 6.2–16.3 percentage points when removed. This highlights the importance of clean, normalized visual inputs for reliable OCR and downstream extraction.

Structured prompting delivers modest overall gains, yet remains valuable in practice. Although the average improvement is limited, it is particularly effective for handling corner cases, which can significantly boost the accuracy of specific fields.

\subsubsection{Document-Type Analysis}
We observe consistent performance differences across document types. Payslips achieve higher accuracy than financial statements in both the baseline and the full pipeline to their inherent structural characteristics. Payslips are shorter and exhibit less layout heterogeneity compared to financial statements. For example, with PaddleOCR and MiniCPM, the full pipeline reaches 96.92\% on payslips compared to 83.95\% on financial statements. The same trend persists across all experimental OCR-VLM settings.  

Notably, the performance gain from the multistage pipeline is substantially greater for financial statements. Accuracy improves by over \textbf{40 percentage points} for financial statements, compared to roughly \textbf{8 percentage points} for payslips. This indicates that the pipeline is particularly beneficial for long and structurally complex financial documents, where retrieval and pre-processing are critical for isolating relevant content and reducing noise.

\section{Error Analysis and Limitations}

We examine common failure cases of the proposed pipeline and summarize its key limitations.

Errors mainly arise from three sources. First, inconsistent financial terminology across documents (e.g., revenue, income, sales) can lead to retrieval mismatches and incorrect field extraction. Second, OCR errors due to low-quality scans, handwriting, or overlapping artifacts may corrupt or omit critical text, in some cases preventing correct page retrieval entirely. Third, currency unit ambiguity in multilingual settings (e.g., IDR’000, ribuan, juta) can result in normalization errors due to magnitude misinterpretation.

The system also has several limitations. Prompt and retrieval query design require per-field manual specification, limiting generalization to new fields. In addition, prompts were primarily optimized for the PaddleOCR + MiniCPM-o-2.6 setting, introducing potential prompt–model alignment bias and reducing transferability across VLM backbones. Finally, performance remains limited for documents with mixed printed and handwritten content.

These observations suggest directions for future work, including terminology normalization, domain-specific multilingual lexicons, improved OCR robustness, and learning-based currency unit disambiguation.
\section{Conclusion}

We presented a multistage framework for structured extraction from long, multilingual scanned financial documents in industrial KYC workflows. Through evaluation on 120 real-world documents, we showed that decoupling page-level retrieval from multimodal reasoning substantially improves extraction accuracy while maintaining comparable service latency. Across multiple OCR–VLM combinations, the proposed pipeline consistently outperforms direct PDF-to-VLM baselines, with gains of up to \textbf{31.9 percentage points}. Our analysis highlights page-level retrieval as the dominant factor in performance improvement and demonstrates that compact VLMs can achieve strong results when supported by appropriate system design. These findings provide practical guidance for building scalable and reliable document extraction systems in regulated financial environments.



\bibliography{custom}

\appendix
\section{Additional Dataset Details}
\label{sec:dataset}

\subsection{Distribution of Document Page}

\begin{figure}[t]
\centering
\includegraphics[width=\linewidth]{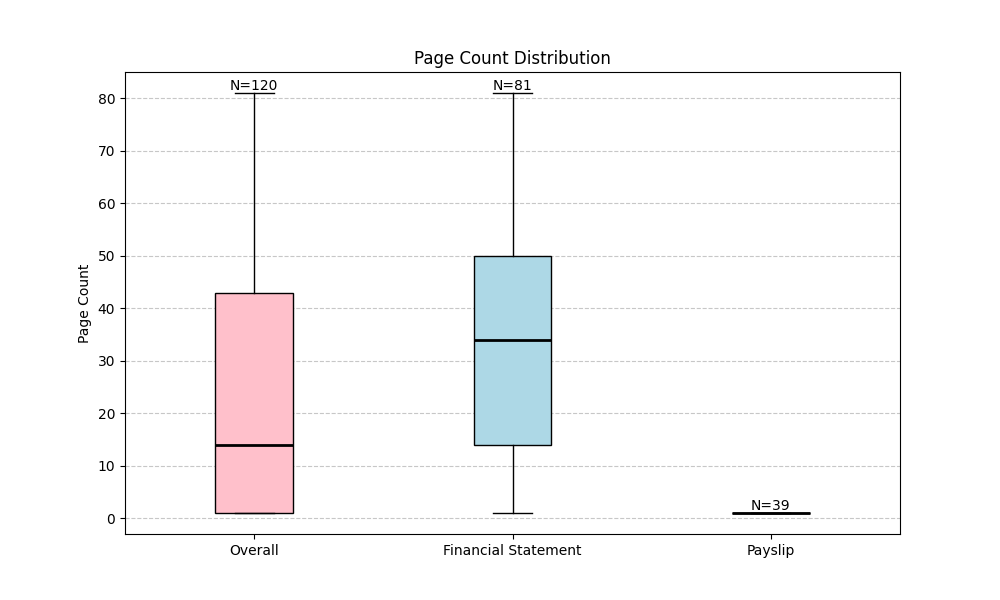}
\caption{Page count distribution across document types. Box plots illustrate the distribution of document length for the overall dataset, financial statements, and payslips. Numbers above each box denote the total number of documents for that category.}
\label{fig:page_distribution}
\end{figure}

Figure \ref{fig:page_distribution} presents the distribution of page counts across document types. Among 120 documents, we have 89 financial statement and 31 payslips. Financial statements are substantially longer ranging from 2 to 81 pages, with a median length exceeding 30 pages and a long right tail. In contrast, payslips are short and highly concentrated, ranging from 1 to 3 pages with average length about 1.5 pages. The overall distribution reflects this mixture, exhibiting strong right-skewness driven by lengthy financial statements, with average document length about 24 pages. This highlights the realistic long-document setting considered in our evaluation, where only a small subset of pages is relevant for downstream extraction.

\subsection{Language Distribution}
Language distribution of documents is shown in Table \ref{tab:language_distribution}

\begin{table*}[t]
\centering
\begin{tabular}{lccc}
\hline
\textbf{Language} & \textbf{Financial Statement} & \textbf{Payslip} & \textbf{Combined} \\
\hline
English & 58 (72\%) & 10 (26\%) & 68 (57\%) \\
Simplified Chinese & 11 (14\%) & 11 (28\%) & 22 (18\%) \\
Bahasa & 7 (9\%) & 11 (28\%) & 18 (15\%) \\
Traditional Chinese & 5 (6\%) & 7 (18\%) & 12 (10\%) \\
\hline
Total & 81 & 39 & 120 \\
\hline
\end{tabular}
\caption{Language distribution across document types.}
\label{tab:language_distribution}
\end{table*}

\subsection{Extraction Field Definition}
To standardize the evaluation process, we defined a fixed schema of target fields specific to the document type. Table \ref{tab:fields} outlines the consolidated list of target fields for Financial Statements and Payslips, categorized by their data type (Text or Numeric). This schema covers 12 distinct data points essential for downstream financial analysis.

\begin{table}[h]
\centering
\small
\begin{tabular}{lll}
\toprule
\textbf{Document Type} & \textbf{Field} & \textbf{Field Type} \\
\midrule
\multirow{7}{*}{Financial Statement} 
& Company Name  & Text \\
& Currency      & Text \\
& Dividend      & Numeric \\
& Total Equity  & Numeric \\
& Net Profit    & Numeric \\
& Revenue       & Numeric \\
& Year          & Numeric \\
\midrule
\multirow{5}{*}{Payslip} 
& Currency Unit & Text \\
& Net Pay       & Numeric \\
& Month         & Text \\
& Year          & Numeric \\
& Commission    & Numeric \\
\bottomrule
\end{tabular}
\caption{Consolidated target fields by document type.}
\label{tab:fields}
\end{table}

\section{Additional Experiment Details}
\label{sec:result}

\subsection{Latency Comparison}
\label{subsec:latency}
\begin{figure*}
  \centering
  \includegraphics[width=\linewidth]{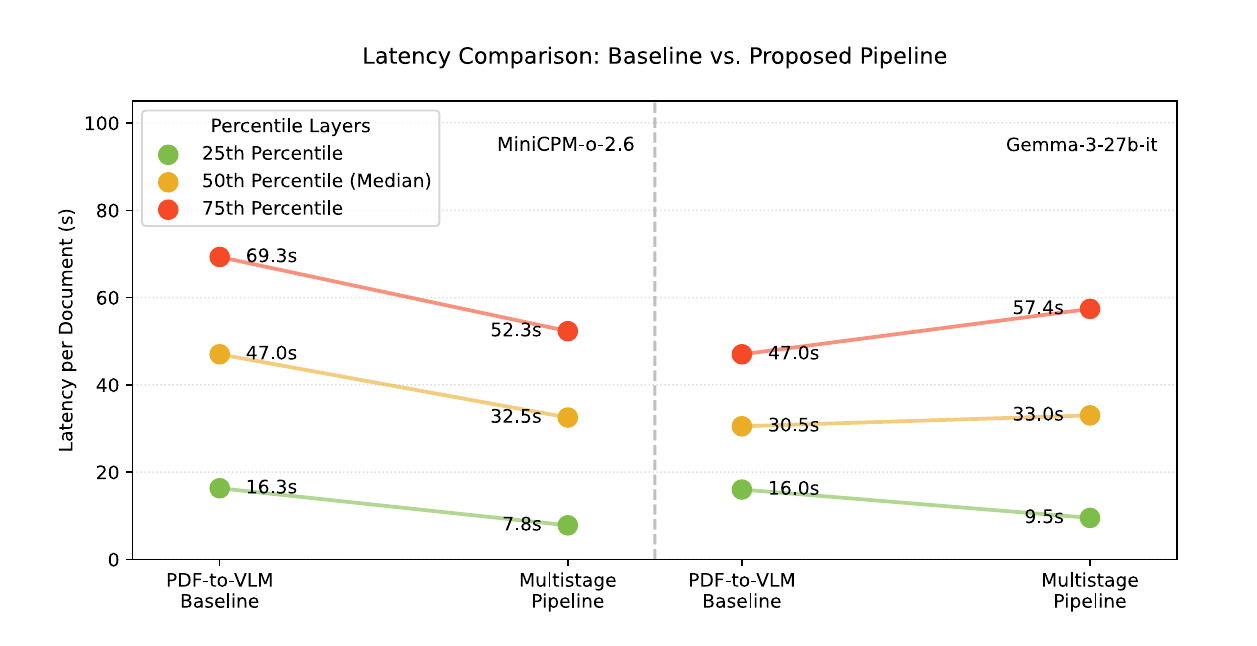}
  \caption{Latency comparison between the direct PDF-to-VLM baseline and the proposed multistage pipeline under PaddleOCR with MiniCPM-o-2.6 and Gemma-3-27B-IT. The plot reports 25th, 50th (median), and 75th percentile document-level latency.}
  \label{fig:latency}
\end{figure*}

As shown in the latency comparison plot Figure \ref{fig:latency}, the per-document latency distributions of the multistage pipeline remain broadly comparable to the direct PDF-to-VLM baseline across both MiniCPM-o-2.6 and Gemma-3-27B-IT. For the PDF-to-VLM baseline, latency statistics are computed only on runs that completed successfully without out-of-memory (OOM) errors. For MiniCPM-o-2.6, latency percentiles decrease consistently under the pipeline setting, while for Gemma-3-27B-IT, median and lower-percentile latency remain similar, with only minor variation at the upper percentile. Overall, these results indicate that the substantial accuracy gains achieved by the proposed framework do not come at the cost of increased latency, supporting its practicality for real-world financial document processing.

\subsection{Visualization of Mudule Contribution}
\label{sec:result_module}
\begin{figure*}[t]
\centering
\includegraphics[width=\linewidth]{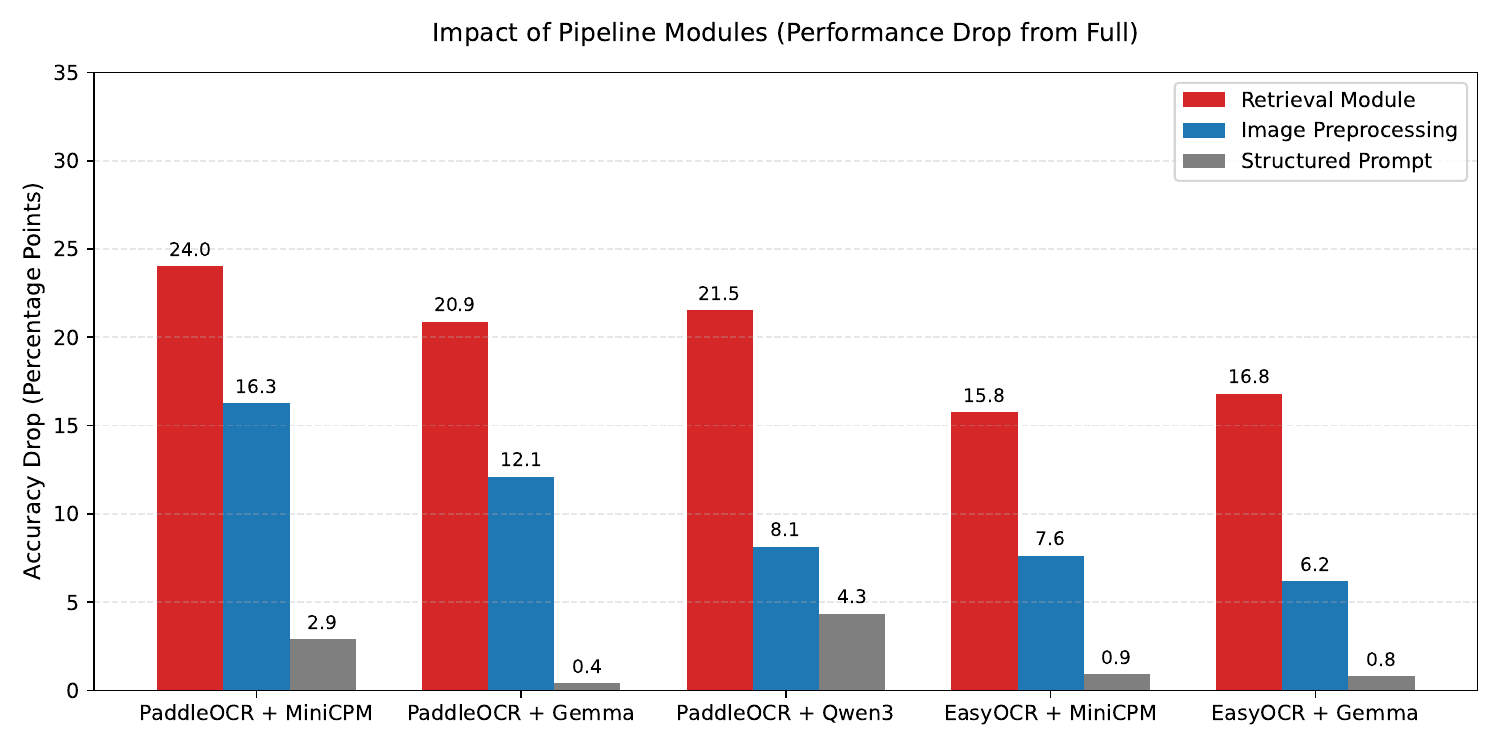}
\caption{Absolute accuracy drop of module removing compared with performance of proposed full pipeline}
\label{fig:ablation}
\end{figure*}

Figure \ref{fig:ablation} quantifies the contribution of each pipeline component by measuring the absolute percentage point drop in accuracy when that specific module is removed. The resulting hierarchy is consistent across all four OCR–VLM configurations, establishing a clear order of architectural priority.

\subsection{Accuracy by Document Type}
We further analyzed how different pipeline components impact performance across specific document categories. Table \ref{tab:doctype_ablation} and Table \ref{tab:doctype_breakdown} present the field-level accuracy for Financial Statements and Payslips under various ablation settings.

\begin{figure*}[t]
\centering
\includegraphics[width=\linewidth]{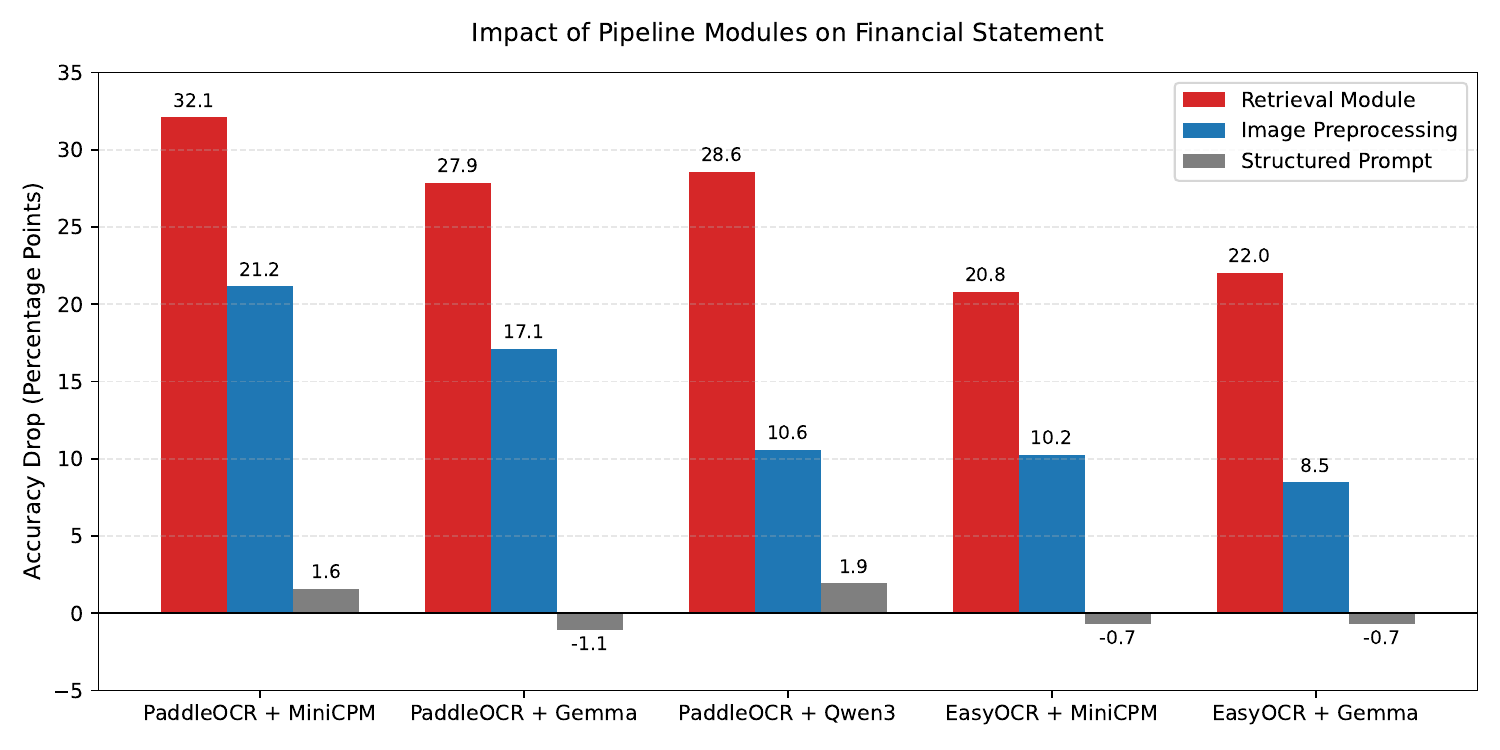}
\caption{Absolute accuracy change for financial statement of module removing compared with performance of proposed full pipeline}
\label{fig:ablation_fin}
\end{figure*}

As illustrated in Figure \ref{fig:ablation_fin}, page retrieval contributes the largest share of the overall accuracy gain, followed by image preprocessing. This trend is consistent with the observations reported in Appendix \ref{sec:result_module}. In contrast, structured prompting yields mixed results. Only the PaddleOCR with MiniCPM-o-2.6 configuration shows a clear positive impact, while other OCR–VLM combinations exhibit marginal declines. One possible explanation is that the human-in-the-loop refinements were primarily developed from the deployed PaddleOCR–MiniCPM setting. As a result, the prompt design may be implicitly tailored to this configuration, leading to better alignment and improved performance for this pair but limited generalization to others.

\begin{figure*}[t]
\centering
\includegraphics[width=\linewidth]{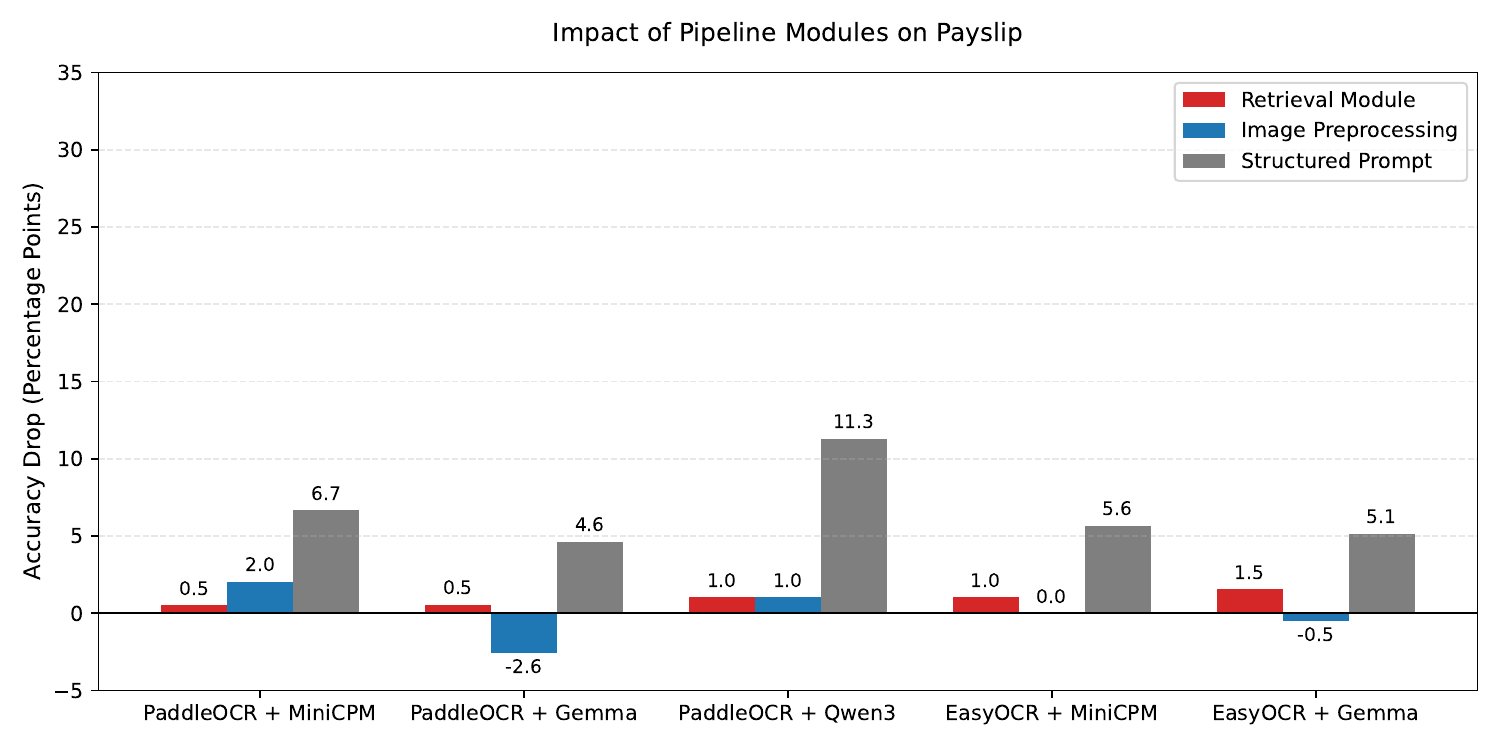}
\caption{Absolute accuracy change for payslip of module removing compared with performance of proposed full pipeline}
\label{fig:ablation_pay}
\end{figure*}

As shown in Figure \ref{fig:ablation_pay}, the module-wise contributions for payslip documents exhibit a markedly different pattern. Structured prompting provides the largest performance gain, while page retrieval and image preprocessing contribute minimally and, in some cases, introduce slight negative effects. A plausible explanation is that payslips are short, visually clean, and relatively standardized in layout. As a result, the direct PDF-to-VLM baseline already achieves strong performance, leaving limited room for improvement from retrieval or preprocessing. In this setting, task-specific prompt design plays a more decisive role in guiding accurate extraction.

\subsection{Accuracy by Language}

The system's adaptability to different languages is quantified in Table \ref{tab:detailed_breakdown}. While performance on English documents remains competitive across most configurations, the specific combination of PaddleOCR and MiniCPM-o-2.6 demonstrates superior robustness on Non-English content, achieving 92.81\% accuracy. This indicates that model choice is particularly critical when processing multilingual financial datasets.

  \begin{table*}[t]
  \centering
  \small
  \setlength{\tabcolsep}{6pt}
  \begin{tabular}{lllccccc}
  \toprule
  \textbf{OCR} & \textbf{VLM} & \textbf{Doc Type} & \textbf{Full} & \textbf{-ImgPrep} & \textbf{-Retr.} &
  \textbf{-Prompt} & \textbf{Direct VLM} \\
  \midrule
  \multirow{4}{*}{PaddleOCR} & \multirow{2}{*}{MiniCPM-o-2.6} & Financial Stmt. & \textbf{83.95\%} & 62.79\% &
  51.85\% & 82.36\% & 43.92\% \\
                             &                                & Payslip         & \textbf{96.92\%} & 94.87\% &
  96.41\% & 90.26\% & 88.72\% \\
  \cmidrule{2-8}
                             & \multirow{2}{*}{Gemma-3-27b-it} & Financial Stmt. & 69.49\% & 52.38\% & 41.62\%
  & \textbf{70.55\%} & 36.33\% \\
                             &                                & Payslip         & 83.08\% & \textbf{85.64\%} &
  82.56\% & 78.46\% & 80.51\% \\
  \cmidrule{2-8}
                             & \multirow{2}{*}{Qwen3-VL-8B-Instruct}& Financial Stmt. & \textbf{82.36\%}& 71.78\%&
  53.79\%& 80.42\%&46.91\%\\
                             &                                & Payslip         & \textbf{93.85\%}& 92.82\%& 92.82\%&
  82.56\%&82.05\%\\
  \midrule
  \multirow{4}{*}{EasyOCR}   & \multirow{2}{*}{MiniCPM-o-2.6} & Financial Stmt. & 70.37\% & 60.14\% & 49.56\% &
   \textbf{71.08\%} & 44.09\% \\
                             &                                & Payslip         & \textbf{89.23\%} &
  \textbf{89.23\%} & 88.21\% & 83.59\% & 83.59\% \\
  \cmidrule{2-8}
                             & \multirow{2}{*}{Gemma-3-27b-it} & Financial Stmt. & 59.96\% & 51.50\% & 37.92\%
  & \textbf{60.67\%} & 37.21\% \\
                             &                                & Payslip         & 80.00\% & \textbf{80.51\%} &
  78.46\% & 74.87\% & 77.44\% \\
  \bottomrule
  \end{tabular}
  \caption{Field-level accuracy comparison across document types and pipeline configurations.}
  \label{tab:doctype_ablation}
  \end{table*}


  \begin{table*}[t]
  \centering
  \small
  \setlength{\tabcolsep}{8pt}
  \begin{tabular}{ll ccc ccc}
  \toprule
  & & \multicolumn{2}{c}{\textbf{Overall}} & \multicolumn{2}{c}{\textbf{Financial Stmt.}} &
  \multicolumn{2}{c}{\textbf{Payslip}} \\
  \cmidrule(lr){3-4} \cmidrule(lr){5-6} \cmidrule(lr){7-8}
  \textbf{OCR} & \textbf{VLM} & \textbf{Count} & \textbf{\%} & \textbf{Count} & \textbf{\%} & \textbf{Count} &
  \textbf{\%} \\
  \midrule
  \multirow{3}{*}{PaddleOCR} & MiniCPM-o-2.6        & 665/762 & 87.27\% & 476/567 & 83.95\% & 189/195 &
  \textbf{96.92\%} \\
                             & Gemma-3-27b          & 556/762 & 72.97\% & 394/567 & 69.49\% & 162/195 &
  \textbf{83.08\%} \\
                             & Qwen3-VL-8B-Instruct & 650/762 & 85.30\% & 467/567 & 82.36\% & 183/195 &
  \textbf{93.85\%} \\
  \midrule
  \multirow{2}{*}{EasyOCR}   & MiniCPM-o-2.6        & 573/762 & 75.20\% & 399/567 & 70.37\% & 174/195 &
  \textbf{89.23\%} \\
                             & Gemma-3-27b          & 496/762 & 65.09\% & 340/567 & 59.96\% & 156/195 &
  \textbf{80.00\%} \\
  \bottomrule
  \end{tabular}
  \caption{Full Pipeline Field-level Extraction Accuracy by Document Type.}
  \label{tab:doctype_breakdown}
  \end{table*}

  \begin{table*}[t]                                                                                            
  \centering
  \small                                                                                                       
  \setlength{\tabcolsep}{8pt}
  \begin{tabular}{ll ccc ccc}
  \toprule
  & & \multicolumn{2}{c}{\textbf{Overall}} & \multicolumn{2}{c}{\textbf{English}} &
  \multicolumn{2}{c}{\textbf{Non-English}} \\
  \cmidrule(lr){3-4} \cmidrule(lr){5-6} \cmidrule(lr){7-8}
  \textbf{OCR} & \textbf{VLM} & \textbf{Count} & \textbf{\%} & \textbf{Count} & \textbf{\%} & \textbf{Count} &
  \textbf{\%} \\
  \midrule
  \multirow{3}{*}{PaddleOCR} & MiniCPM-o-2.6        & 665/762 & 87.27\% & 381/456 & 83.55\% & 284/306 &
  \textbf{92.81\%} \\
                             & Gemma-3-27b          & 556/762 & 72.97\% & 338/456 & \textbf{74.12\%} & 218/306
  & 71.24\% \\
                             & Qwen3-VL-8B-Instruct & 650/762 & 85.30\% & 366/456 & 80.26\% & 284/306 &
  \textbf{92.81\%} \\
  \midrule
  \multirow{2}{*}{EasyOCR}   & MiniCPM-o-2.6        & 573/762 & 75.20\% & 365/456 & \textbf{80.04\%} & 208/306
  & 67.97\% \\
                             & Gemma-3-27b          & 496/762 & 65.09\% & 314/456 & \textbf{68.86\%} & 182/306
  & 59.48\% \\
  \bottomrule
  \end{tabular}
  \caption{Full Pipeline Field-level Extraction Accuracy by Language.}
  \label{tab:detailed_breakdown}
  \end{table*}

\section{Process Visualization}
\begin{figure*}
  \centering
  \includegraphics[width=\linewidth]{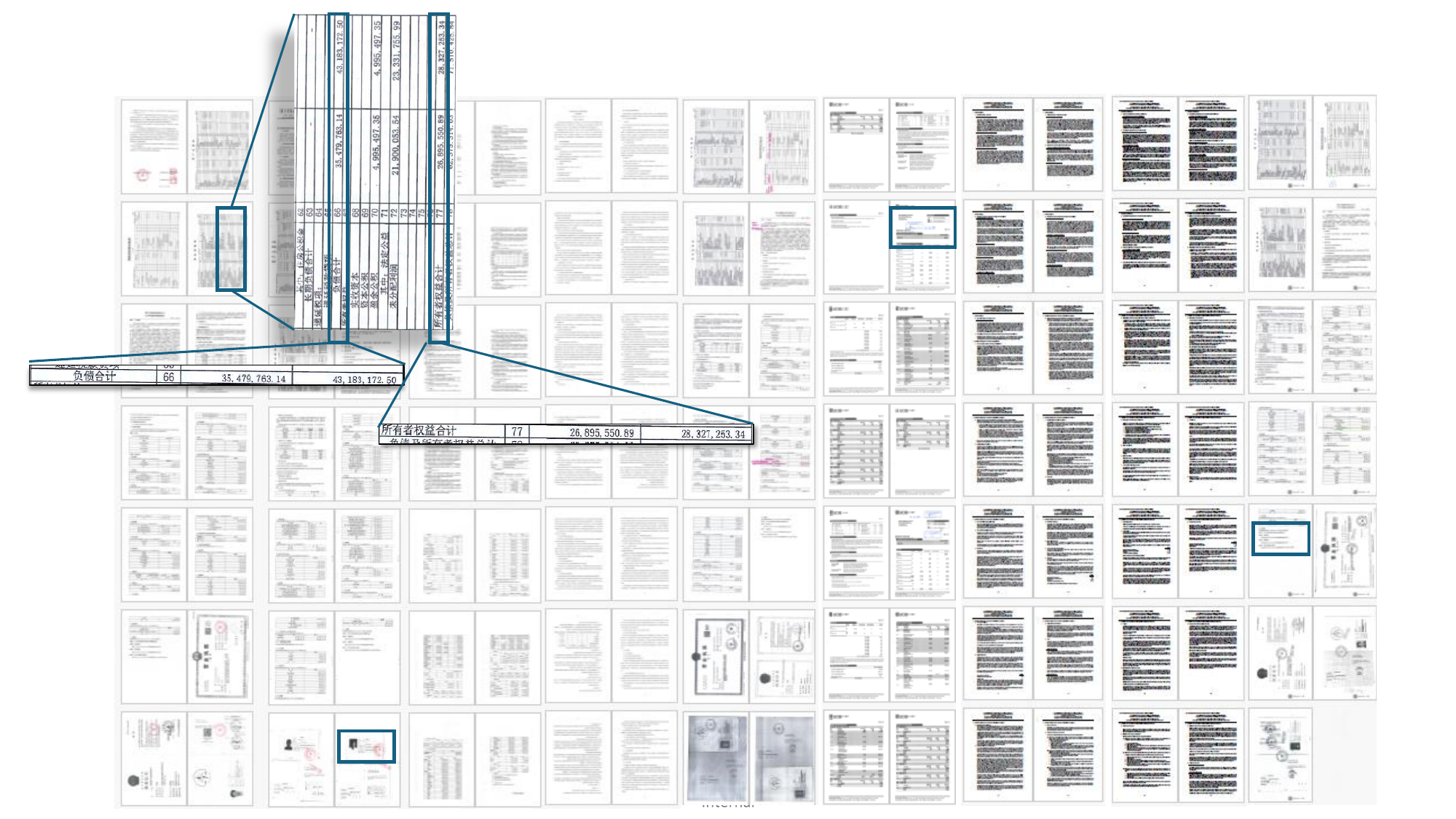}
  \caption{Visualization of Information Sparsity and Preprocessing Requirements in Long Financial Document}
  \label{fig:overall_process}
\end{figure*}

\begin{figure*}
  \centering
  \includegraphics[width=\linewidth]{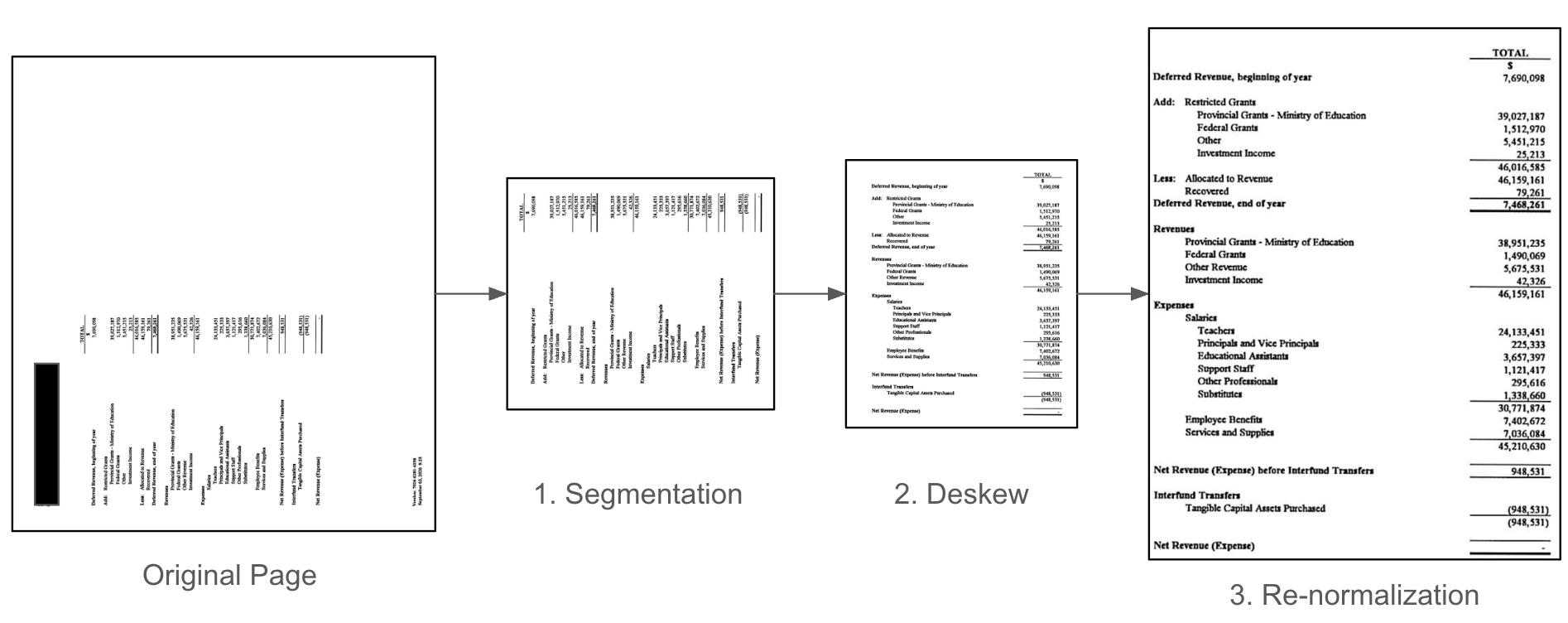}
  \caption{The illustration of image pre-processing. It consists of three steps, including page segmentation, deskew, and re-normailzation. These steps are critical to the performance of the downstream OCR and information extraction.}
  \label{fig:preprocessing}
\end{figure*}

\begin{figure*}
  \centering
  \includegraphics[width=\linewidth]{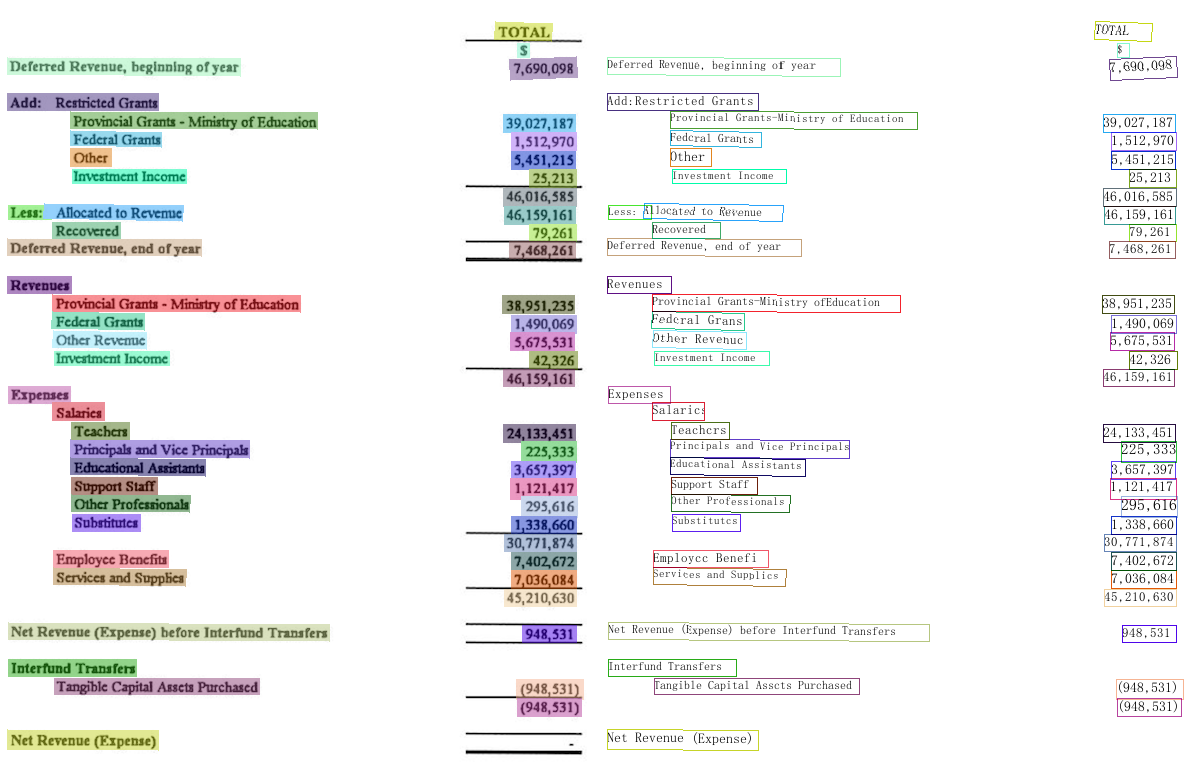}
  \caption{Sample OCR transcription output showing recognized text and bounding box coordinates for layout-aware extraction.}
  \label{fig:ocr}
\end{figure*}

This section provides some visualisations of how the real-world Know-Your-Customer (KYC) document is processed by the proposed pipeline. Figure \ref{fig:overall_process} illustrate the extraction process from multipage real world financial document. It shows industrial financial documents exhibit extreme information sparsity, where a single target field—such as net profit or revenue—may be located on only one or two pages within a document exceeding 50 pages. 

Figure \ref{fig:preprocessing} illustrates the image pre-processing steps. It consists of three steps, including page segmentation, deskew,
and re-normailzation. And Figure \ref{fig:ocr} shows the sample output from OCR stage.

\section{Example VLM Prompt}
\label{sec:query}

To illustrate how extraction prompts are constructed in our framework, we present an example of extracting Dividend from English financial statements as shown below:






{\small
\begin{Verbatim}[breaklines=true, breakanywhere=true]
You are an expert financial data extraction
specialist.

Extract the dividend from the given document.

Key financial terms: dividend, paid, financial,
statement

Do not extract: Dividends declared or approved,
but not yet paid.
Do not extract: Dividends received by the
company.
Do not extract: Dividend amount per share.
Do not extract: Adjustments for dividend income.
Do not extract: Stock dividends.

Output result in JSON only. Do NOT change JSON
key. Return a list for data with multiple
years.
Leave blank empty if unsure, and specify reason
in key remarks
\end{Verbatim}
}



\end{document}